\documentclass[letterpaper, 10 pt, conference]{ieeeconf}  % Comment this line out
\usepackage[utf8x]{inputenc}

\usepackage{amsmath}
\usepackage{amssymb}
\usepackage{multicol}
\usepackage{booktabs}
\usepackage{graphicx}
\usepackage{soul}
\usepackage{xcolor}
\usepackage{rotating}
\usepackage{booktabs}

\IEEEoverridecommandlockouts                              % This command is only
\newcommand\floor[1]{\lfloor#1\rfloor}

\usepackage[normalem]{ulem} % for \sout

\title{\LARGE \bf
RIU-Net: Embarrassingly simple semantic segmentation\\of 3D LiDAR point cloud
}

\author{Pierre Biasutti$^{1,2,3}$, Aurélie Bugeau$^{1}$, Jean-François Aujol$^{2}$, Mathieu Brédif$^{3}$\\
$^1$\textit{Univ. Bordeaux, CNRS,  Bordeaux INP, LaBRI, UMR 5800, F-33400, Talence, France}\\
$^2$\textit{Univ. Bordeaux, IMB, INP, CNRS, UMR 5251, F-33400 Talence, France}\\
$^3$\textit{Univ. Paris-Est, LASTIG GEOVIS, IGN, ENSG, F-94160 Saint-Mandé, France}\\
pierre.biasutti@labri.fr}

\begin{document}

\maketitle
\thispagestyle{empty}
\pagestyle{empty}

%%%%%%%%%%%%%%%%%%%%%%%%%%%%%%%%%%%%%%%%%%%%%%%%%%%%%%%%%%%%%%%%%%%%%%%%%%%%%%%%
\begin{abstract}
This paper proposes RIU-Net (for Range-Image U-Net), the adaptation of a popular semantic segmentation network for the semantic segmentation of a 3D LiDAR point cloud. The point cloud is turned into a 2D range-image by exploiting the topology of the sensor. This image is then used as input to a U-net. This architecture has already proved its efficiency for the task of semantic segmentation of medical images. We demonstrate how it can also be used for the accurate semantic segmentation of a 3D LiDAR point cloud and how it represents a valid bridge between image processing and 3D point cloud processing. Our model is trained on range-images built from KITTI 3D object detection dataset. Experiments show that RIU-Net, despite being very simple, offers results that are comparable to the state-of-the-art of range-image based methods. Finally, we demonstrate that this architecture is able to operate at 90fps on a single GPU, which enables deployment for real-time segmentation.

\end{abstract}

%%%%%%%%%%%%%%%%%%%%%%%%%%%%%%%%%%%%%%%%%%%%%%%%%%%%%%%%%%%%%%%%%%%%%%%%%%%%%%%%
\section{Introduction}
The recent interest for autonomous systems have motivated many computer vision works over the past years. Indeed, the importance of accurate perception models is a crucial step towards systems automation, especially for mobile robots and autonomous driving. Modern systems are equipped with both optical cameras and 3D sensors (namely LiDAR sensors). LiDAR sensors are essential components of modern perception systems as they enable direct space measurements, providing accurate 3D representation of the scene. However, for most automation-related tasks, the raw LiDAR point cloud needs further processing in order to be used. In particular, point cloud with accurate semantic segmentation provides a higher level of representation of the scene that can be used in various applications such as obstacle avoiding, road inventor or object manipulation.

\begin{figure}
    \centering
    \newcommand{\szw}{0.45}
    \begin{tabular}{r@{ }c}
        \begin{turn}{90}\tiny{\quad prediction}\end{turn} & \includegraphics[width=\szw\textwidth]{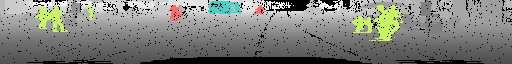} \\
        \begin{turn}{90}\tiny{\; groundtruth}\end{turn} & \includegraphics[width=\szw\textwidth]{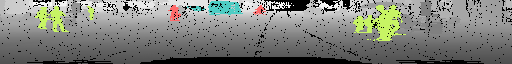} \\
        \begin{turn}{90}\tiny{\quad \quad \quad \quad \quad prediction}\end{turn} &\includegraphics[width=\szw\textwidth]{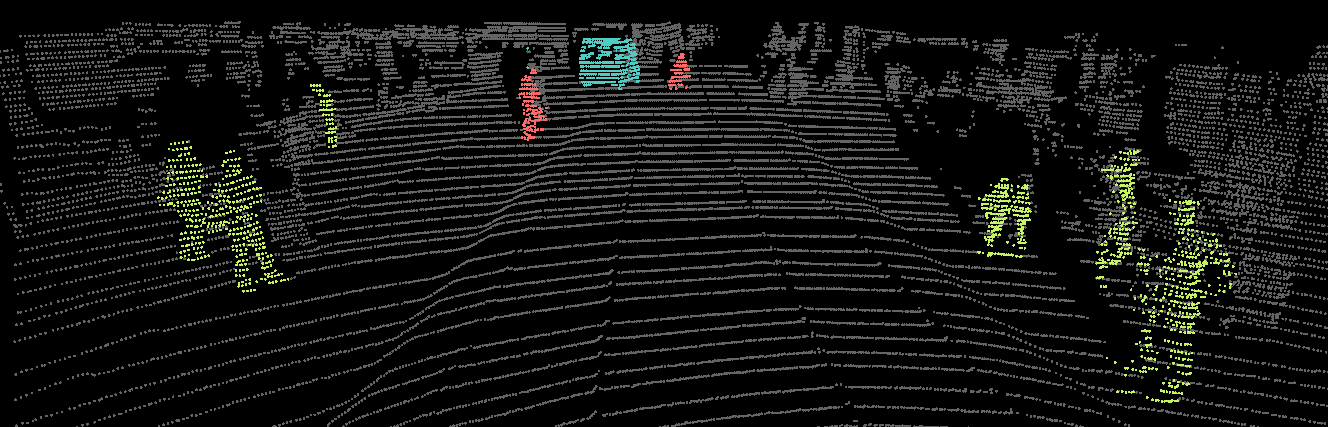} \\
        \begin{turn}{90}\tiny{\quad \quad \quad \quad \quad groundtruth}\end{turn} &\includegraphics[width=\szw\textwidth]{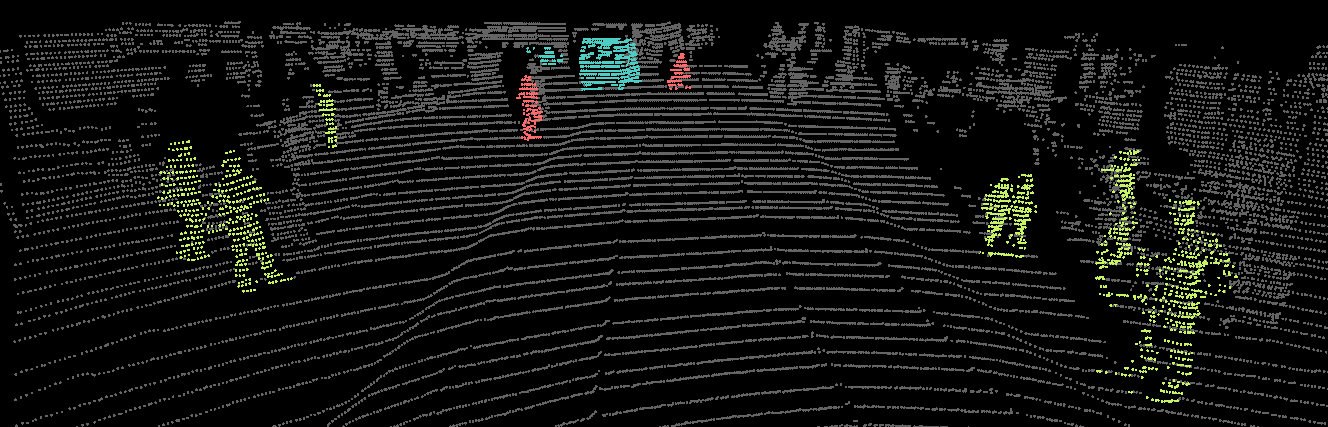} \\
    \end{tabular}
    \caption{Result of the range-image semantic segmentation produced by the proposed method. The first two results show the prediction of the proposed model and the groundtruth respectively, seen in the sensor topology. The last two results show the same prediction and groundtruth in 3D. }
    \label{fig:front}
\end{figure}

This work focuses on semantic segmentation of 3D LiDAR point clouds. Given a point cloud acquired with a LiDAR sensor, we aim at estimating a label for each point that belongs to objects of interests in urban environment (such as cars, pedestrians and cyclists). The traditional pipelines used to tackle this problem consider ground removal, clustering of remaining structures, and classification based on handcrafted features extracted on each clusters \cite{himmelsbach2008lidar,feng2014fast}. The segmentation can be improved with variational models \cite{landrieu2018large}. These methods are often hard to tune as handcrafted features usually require tuning many parameters, which is likely to be data dependant and therefore hard to use in a general scenario. Finally, although the use of regularization can lead to visual and qualitative improvements, it often leads to a large increasing of the computational time.

Recently, deep-learning approaches have been proposed to overcome the problem of the difficulty of tuning handcrafted features. This has become possible with the arrival of large 3D annotated datasets such as the KITTI 3D object detection dataset \cite{geiger2012are}. Many methods have been proposed to segment the point cloud by directly operating in 3D \cite{qi2017pointnet} or on voxel-based representation of the point cloud \cite{zhou2018voxelnet}. However, this type of methods either need very high computational power, or are not able to process a point cloud that corresponds to a full turn of the sensor in a single pass. Recently, faster approaches have been proposed \cite{wu2018squeezeseg,wang2018pointseg}. They rely on a 2D representation of the point cloud, called range-image, that can be used as the input of a convolutional neural network. Thus, the processing time as well as the required computational power can be kept low. %\ab{\st{Unfortunately, these systems have not yet achieved good enough scores for practical use, especially on small objects classes such as cyclists or pedestrians.}}%that are practically useful on certains classes. 

In this paper, we propose RIU-Net (for Range-Image U-Net), the adaptation of U-Net \cite{ronneberger2015u}, a very popular semantic segmentation architecture, to the semantic segmentation of 3D LiDAR point clouds. We demonstrate that, beside being a very simple architecture, the results of RIU-Net are comparable state-of-the-art range-image methods, as shown in Figure \ref{fig:front}.% \st{We demonstrate that, beside being a straightforward adaptation, the results of RIU-Net outperform state-of-the-art range-image methods, as shown in Figure \ref{fig:front}. We also propose a lighter version of the network which requires as low memory as state-of-the-art methods. Finally, both methods require similar, if not lower, computational time.}

The contributions of the paper are the followings: 1) a simple adaptation of the method presented in  \cite{ronneberger2015u} for the accurate semantic segmentation of 3D LiDAR point cloud, 2) a comparison with state-of-the-art methods in which we show that RIU-Net performs better on the same training set. The paper is organized as follows: first, previous works on point cloud semantic segmentation are presented. After that, the details of the adaptation of the model presented in \cite{ronneberger2015u} are explained. Finally, qualitative and quantitative results are shown and a conclusion is drawn.

\section{Related works}

In this section, previous works on image semantic segmentation as well as 3D point cloud semantic segmentation are presented.

\subsection{Semantic segmentation for images}

Semantic segmentation of images has been the subject of many works in the past years. Recently, deep-learning methods have largely outperformed existing methods. The method presented in \cite{long2015fully} was the first to propose an accurate end-to-end network for semantic segmentation. This method is based on an encoder in which each scale is used to compute the final segmentation. Only a few month later, the U-net architecture \cite{ronneberger2015u} (later generalized in \cite{badrinarayanan2017segnet}) has been proposed for the semantic segmentation of medical images. This method is an encoder-decoder that is able to reach very fine precision in the segmentation. These two methods have largely influenced recent works such as DeeplabV3+ \cite{chen2018deeplabv3plus} that uses dilated convolutional layers and spatial pyramid pooling modules in an encoder-decoder structure to improve the quality of the prediction. Other approaches explore multi-scale architectures to produce and fuse segmentations performed at different scales \cite{lin2017refinenet,zhao2018icnet}. Most of these methods are able to produce very accurate results, on various types of images (medical, outdoor, indoor). % However, they often require heavy computational power and do not operate in real-time \ab{Je ne sais pas trop si on peut vraiment dire ça ici}.  
The review \cite{briot2018analysis} of CNNs methods for semantic segmentation provides a deep analysis of some recent techniques. This work demonstrates that a combination of various components would most likely improve segmentation results on wider classes of objects. 
%Nevertheless, the approaches presented in U-net \cite{ronneberger2015u} and \cite{zhao2018icnet} have demonstrated that they can operate in real-time on low resolution images.
%\ab{tu veux vraiment parler du temps de calcul ? - En fait quelle est le message à faire passer ici. Pour moi il faut dire que y a plein de méthodes qui marchent déjà bien pour les images. On pourra ensuite dans la conclusion sire que si on rajoute à unet des composants des architecutres plus ocmpliquees de segmentation semantique d'images, on pourra surement ameliorer les resultats. Qu'en penses tu ?
%-> Oui clairement le message a faire passer c'est que 1) l'état de l'art niveau image est bien fourni, 2) ça marche très bien, 3) c'est raisonnable pour de l'embarqué, mais c'est pas forcément utile de le mettre ici, puisque c'est aussi le cas des méthodes basées range-image.}

\subsection{Semantic segmentation for point clouds}

\paragraph{3D-based methods} As mentioned above, the first approaches for point cloud semantic segmentation were done using heavy pipelines, composed of many successive steps such as: ground removal, point cloud clustering, feature extraction as presented in \cite{himmelsbach2008lidar,feng2014fast}. 
%A final regularization step was also proposed to improve the coherence of the segmentation \cite{landrieu2018large} \mb{je presenterai plutot celui ci comme une approche type super pixel, la tournure de la phrase suivante semble impliquer, a tort, que celui ci n'est pas du deep...}\ab{Est ce que tu parles du bon papier en fait là ?}\ab{Another approach~\cite{landrieu2018large}  proposes superpoint graph as a representation of the point cloud. The graph is then segmented with  a  graph  convolutional network}. 
However, these methods often require many parameters and they are therefore hard to tune. Recently, Landrieu et al. proposed in \cite{landrieu2019point} to extract features of the point cloud using a deep-learning approach. Then, the segmentation is done using a variational regularization. Another approach presented in \cite{qi2017pointnet} proposes to directly input the raw 3D LiDAR point cloud to a network composed of a succession of fully-connected layers to classify or segment the point cloud. However, due to the heavy structure of this architecture, it is only suitable for small point clouds. Moreover, processing 3D data often increases the computational time due to the dimension of the data (number of points, number of voxels), and the absence of spatial correlation. To overcome these limitations, the methods presented in \cite{li20173d} and \cite{zhou2018voxelnet} propose to represent the point cloud as a voxel-grid which can be used as the input of 3D CNN. These methods achieve satisfying results for 3D detection. However, semantic segmentation would require a voxel-grid of very high resolution, which would increase the computational cost as well as the memory usage. 

\paragraph{Range-image based methods} Recently, Wu et al. proposed SqueezeSeg \cite{wu2018squeezeseg}, a novel approach for the semantic segmentation of a LiDAR point cloud represented as a spherical range-image \cite{biasutti2018range}. This representation allows to perform the segmentation by using simple 2D convolutions, which lowers the computational cost while keeping good accuracy. The architecture is derived from the SqueezeNet image segmentation method \cite{iandola2016squeezenet}. The intermediate layers are "fire layers", \textit{i.e.} layers made of one squeeze module and one expansion module. 
%where two convolutions of different size are done on the same input and concatenated as an input.  
Later on, the same authors improved this method in \cite{wu2018squeezesegv2} by adding a context aggregation module and by considering focal loss and batch normalization to improve the quality of the segmentation. 
A similar range-image approach was proposed in \cite{wang2018pointseg}, where a Atrous Spatial Pyramid Pooling~\cite{chen2018deeplab} and squeeze  reweighting  layer~\cite{hu2018squeeze} are added. %Here again batch normalization and focal loss are considered. 
%\textit{e.g.} layers where two convolutions of different size are done on the same input and concatenate as an input. 
These range-image methods succeed in real-time computation but require heavy architectures including post-processing steps. %However, their results often lack of accuracy which limits their usage in real scenarios.

In next section, we propose RIU-Net: an adaptation of U-net \cite{ronneberger2015u} for the semantic segmentation of point clouds represented as range-images.

\section{Methodology}

In this section, we present RIU-Net, our adaptation of the U-net architecture \cite{ronneberger2015u} for the semantic segmentation of LiDAR point clouds. The method consists in feeding the U-net architectures with 2-channels images encoding range and elevation. In the next sub-section, we explain how to build these images, called range-images, that were introduced in \cite{biasutti2018range}.

\subsection{Input of the network}
As mentioned above, processing raw LiDAR point clouds is computationally expensive. Indeed, these 3D point clouds are stored as unorganized lists of $(x,y,z)$ Cartesian coordinates. Therefore processing such data, or turning them into voxels involve heavy memory costs. Modern LiDAR sensors often acquire 3D points, with a sampling pattern with a few number of scan lines and quasi uniform angular steps between samples, from which we can build a dense image \cite{biasutti2018range}, the so-called range-image. Indeed each point is defined by two angles and a depth, $(\theta, \phi, d)$ respectively, with steps of ($\Delta\theta, \Delta\phi$) between two consecutive positions. Each point $p$ of the LiDAR point cloud can be mapped to the coordinates $(x,y)$ with $x = \floor{\frac{\theta}{\Delta \theta}}, y = \floor{\frac{\phi}{\Delta \phi}}$ of a 2D range-image, where each channel represents a modality of the measured point. Note that such processing is not required whenever the raw LiDAR data (with beam number) are available. For the rest of this work, we use a range-image named $u$ of $512 \times 64$px with two channels: the depth towards the sensor and the elevation. In perfect conditions, the resulting image is completely dense, without any missing data. However, due to the nature of the acquisition, some measurements are considered invalid by the sensor and they lead to empty pixels (no-data). This happens when the laser beam is highly deviated (\textit{e.g.} when going through a transparent material) or when it does not create any echo (\textit{e.g.} when the beam points in the sky direction). We propose to identify such pixels using a binary mask $m$ equal to $0$ for empty pixels and to $1$ otherwise. The analysis of multi-echo LiDAR scans is subject to future work. 

\subsection{Architecture}

\begin{figure}
    \centering
    \includegraphics[width=0.45\textwidth]{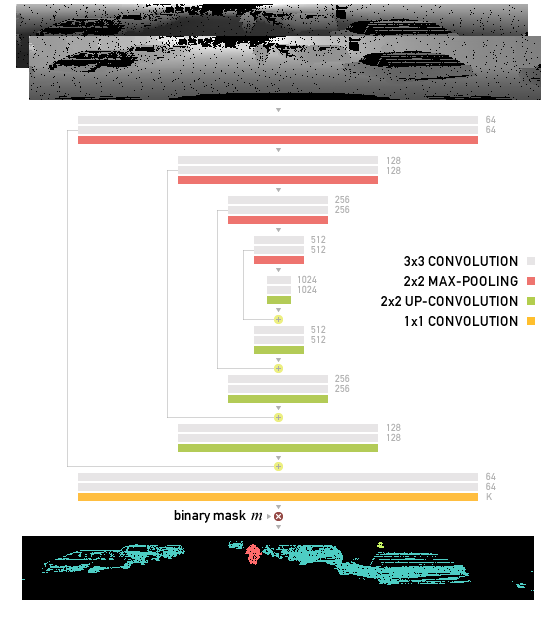}
    \caption{RIU-Net: U-Net architecture adapted to point cloud semantic segmentation with the depth and elevation channels input (top) and the output segmented image (bottom).}
    \label{fig:unet_architecture}
\end{figure}

The U-net architecture \cite{ronneberger2015u} is an encoder-decoder. As illustrated in Figure \ref{fig:unet_architecture}, the first half consists in the repeated application of two $3\times3$ convolutions followed by a rectified linear unit (ReLU) and a $2\times2$ max-pooling layer that downsamples the input by a factor 2. Each time a downsampling is done, the number of features is doubled. The second half of the network consists in upsampling blocks where the input is upsampled using $2\times2$ up-convolutions. Then, concatenation is done between the upsampled feature map and the corresponding feature map of the first half. This allows the network to capture global details while keeping fine details. After that, two $3\times3$ convolutions are applied followed by a ReLU. This block is repeated until the output of the network matches the dimension of the input. Finally, the last layer consists in a 1x1 convolution that outputs as many features as the wanted number of possible labels $K$ 1-hot encoded.

\subsection{Loss function}

The loss function of the semantic segmentation network is defined as the cross-entropy of the softmax of the output of the network. The softmax is defined pixel-wise for each label $k$ as follows: 
\begin{align*}
p_k(x) = \frac{\textrm{exp}(a_k(x))}{\sum\limits_{k'=0}^{K}\textrm{exp}(a_{k'}(x))}
\end{align*}
where $a_k(x)$ is the activation for feature $k$ at the pixel position $x$. After that, we define $l(x)$ the groundtruth label of the $x$ pixel. We then compute the cross-correlation as follows:
\begin{align*}
    E = \sum\limits_{x \in \Omega}\mathbb{1}_{\{m(x) > 0\}}w(x)\textrm{log}(p_{l(x)}(x))
\end{align*}
where $\Omega$ is the domain of definition of $u$, $m(x)>0$ are the valid pixels and $w(x)$ is a weighting function introduced to give more importance to pixels that are close to a separation between two labels, as defined in \cite{ronneberger2015u}.

\subsection{Training}

We train the network with the Adam stochastic gradient optimizer and a learning rate set to $0.001$. We also use batch normalization with a momentum of 0.99 to ensure good convergence of the model.  Finally, the batch size is set to $8$ and the training is stopped after $10$ epochs.

\section{Experiments}

To test RIU-Net, we follow the experimental setup of the SqueezeSeg approach \cite{wu2018squeezeseg} for both training and evaluation. Indeed, they provide range-images with segmentation labels exported from the 3D object detection challenge of the KITTI dataset \cite{geiger2012are}. They also provide the training / validation split that they used for their experiments, which contains $8057$ samples for training and $2791$ for validation.

\begin{figure*}
    \centering
    \newcommand{\szw}{0.48\textwidth}
    \hspace{-0.3cm}\begin{tabular}{cc}
        \includegraphics[width=\szw]{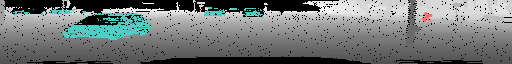} & 
        \includegraphics[width=\szw]{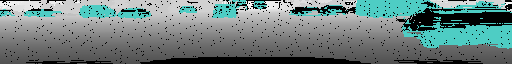} \\
        \includegraphics[width=\szw]{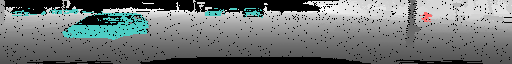} & 
        \includegraphics[width=\szw]{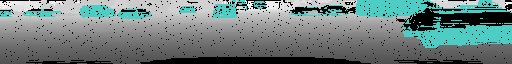} \\
        (a) & (b) \\
        \includegraphics[width=\szw]{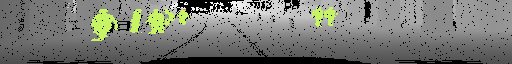} & 
        \includegraphics[width=\szw]{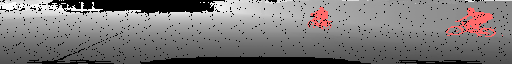} \\
        \includegraphics[width=\szw]{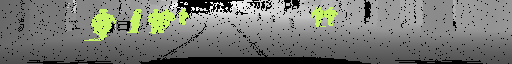} & 
        \includegraphics[width=\szw]{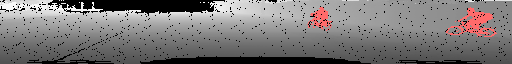} \\
        (c) & (d) \\
    \end{tabular}
    \caption{Results of the semantic segmentation of the proposed method (top) and groundtruth (bottom). Labels are associated to colors as follows: blue for the cars, red for the cyclists and lime for the pedestrians.}
    \label{fig:results}
\end{figure*}

Figure \ref{fig:front} shows a segmentation result of RIU-Net and the groundtruth both on the range-image (top) and in 3D (bottom). The segmentation in 3D is obtained by labelling the raw point cloud according to the result on the range-image. More results are shown in Figure \ref{fig:results}. They all highlight how visually similar the results obtained with RIU-Net and the groundtruth are. %In particular, we can see that the \ab{A reprendre: produced segmentations do not overflow, which is a typical issue with state-of-the-art methods. The quality of the results remains high for every type of object, even on cyclists and pedestrians (Figure \ref{fig:results} (c) and (d)) although they appear about $10$ times less than cars in the training dataset.}

Similarly to \cite{wu2018squeezeseg} and \cite{wu2018squeezesegv2}, we use the intersection-over-union metric to evaluate RIU-Net and we compare it with the state-of-the-art: 
\begin{align*}
    IoU_l = \frac{|\rho_l \bigcap G_l|}{|\rho_l \bigcup G_l|}
\end{align*}
where $\rho_l$ and $G_l$ denote the predicted and groundtruth sets of points that belongs to label $l$ respectively. 

Table \ref{tab:iou} presents the results obtained for the segmentation of cars, cyclists and pedestrians, with SqueezeSeg \cite{wu2018squeezeseg},  SqueezeSegv2 \cite{wu2018squeezesegv2} and PointSeg \cite{wang2018pointseg} compared to RIU-Net. The scores of state-of-the-art methods are taken from the corresponding papers, using the same training conditions. We can see that the results of RIU-Net are comparable to the one of the state-of-the-art, despite the architecture being very simple. Indeed, our method achieves better average IoU scores compared to the PointSeg and the SqueezeSeg architectures. Moreover, it outperforms all the compared methods for cyclists. We believe that the increased number of parameters of our model compensate with the fact that the architecture had not been specifically designed for LiDAR point cloud segmentation in sensor topology, contrary to the other methods. The method SqueezeSegV2 is built on the SqueezeSeg architecture, while proposing several modifications (content aggretation modules and fire layers) driven by the specificity of the data. Therefore, it is reasonable to think that comparable results could be achieved with our model by applying the same modifications, however our goal is to keep the architecture simple and as generic as possible. Visual results of our method against the ground truth are displayed in Figure \ref{fig:results}. Finally, we advocate that the proposed model can operate with a frame-rate of $90$ frames per second on a single GPU, which is comparable, if not faster, to state-of-the-art methods.

\begin{table}[h!]
    \caption{Comparison (IoUs, $\%$) of our approach with the state-of-the-art for the semantic segmentation of the KITTI dataset.}
    \centering
    \begin{tabular}{ccccc}
        \toprule
            &Cars & Pedestrians & Cyclists & Average\\
        \toprule
        PointSeg \cite{wang2018pointseg}      & 67.4 & 19.2 & 32.7 & 39.8\\
        SqueezeSeg \cite{wu2018squeezeseg}    & 64.6 & 21.8 & 25.1 & 37.2\\
        SqueezeSegv2 \cite{wu2018squeezesegv2} & \textbf{73.2}& \textbf{27.8} & {33.6} & \textbf{44.9}\\
        \midrule
        RIU-Net & {62.5} & {22.5} & \textbf{36.8} & {40.6}\\
        \bottomrule
    \end{tabular}
    \label{tab:iou}
\end{table}

\section{Conclusion}
In this paper, we have shown that applying a U-net architecture on LiDAR data is comptetitive with the state-of-the-art for point cloud semantic segmentation. The U-net architectures is fed with range-image representations of a 3D point cloud. The resulting method, called RIU-Net, is simple and fast, but yet provides results that are comparable to state-of-the-art methods, while proposing an efficient bridge between image processing and 3D point cloud processing.
The current method relies on a cross-entropy loss function. We plan to investigate other functions such as the focal loss used in \cite{wu2018squeezesegv2} and to study possible spatial regularization schemes. Finally fusion of LiDAR and optical data would probably enable reaching a higher level of accuracy.

\section{Acknowledgement}
This project has also received funding from the European Union’s Horizon 2020 research and innovation programme under the Marie Skłodowska-Curie grant agreement No 777826.
\bibliographystyle{plain}

\begin{thebibliography}{10}

\bibitem{badrinarayanan2017segnet}
V.~Badrinarayanan, A.~Kendall, and R.~Cipolla.
\newblock Segnet: A deep convolutional encoder-decoder architecture for image
  segmentation.
\newblock {\em IEEE Trans. on Pattern Analysis and Machine Intelligence},
  39(12):2481--2495, 2017.

\bibitem{biasutti2018range}
P.~Biasutti, J-F. Aujol, M.~Br{\'e}dif, and A.~Bugeau.
\newblock {Range-Image: Incorporating sensor topology for LiDAR point cloud
  processing}.
\newblock {\em Photogrammetric Engineering \& Remote Sensing}, 84(6):367--375,
  2018.

\bibitem{briot2018analysis}
A.~Briot, P.~Viswanath, and S.~Yogamani.
\newblock Analysis of efficient {CNN} design techniques for semantic
  segmentation.
\newblock In {\em IEEE Conf. on Computer Vision and Pattern Recognition}, pages
  663--672, 2018.

\bibitem{chen2018deeplab}
L-C. Chen, G.~Papandreou, I.~Kokkinos, K.~Murphy, and A.~L. Yuille.
\newblock Deeplab: Semantic image segmentation with deep convolutional nets,
  atrous convolution, and fully connected crfs.
\newblock {\em IEEE Trans. on Pattern Analysis and Machine Intelligence},
  40(4):834--848, 2018.

\bibitem{chen2018deeplabv3plus}
L-C. Chen, Y.~Zhu, G.~Papandreou, F.~Schroff, and A.~Hartwig.
\newblock Encoder-decoder with atrous separable convolution for semantic image
  segmentation.
\newblock In {\em Proc. of ECCV}, pages 801--808, 2018.

\bibitem{feng2014fast}
C.~Feng, Y.~Taguchi, and V.~R. Kamat.
\newblock Fast plane extraction in organized point clouds using agglomerative
  hierarchical clustering.
\newblock In {\em IEEE International Conference on Robotics and Automation},
  pages 6218--6225, 2014.

\bibitem{geiger2012are}
A.~Geiger, P.~Lenz, and R.~Urtasun.
\newblock {Are we ready for Autonomous Driving? The KITTI Vision Benchmark
  Suite}.
\newblock In {\em IEEE Conf. on Computer Vision and Pattern Recognition}, pages
  3354--3361, 2012.

\bibitem{himmelsbach2008lidar}
M.~Himmelsbach, A.~Mueller, T.~L{\"u}ttel, and H-J. W{\"u}nsche.
\newblock {LIDAR-based 3D object perception}.
\newblock In {\em Proceedings of international workshop on Cognition for
  Technical Systems}, pages 1--7, 2008.

\bibitem{hu2018squeeze}
J.~Hu, L.~Shen, and G.~Sun.
\newblock Squeeze-and-excitation networks.
\newblock In {\em IEEE Conf. on Computer Vision and Pattern Recognition}, pages
  7132--7141, 2018.

\bibitem{iandola2016squeezenet}
F.~N. Iandola, S.~Han, M.~W. Moskewicz, K.~Ashraf, W.~J. Dally, and K.~Keutzer.
\newblock Squeezenet: {AlexNet}-level accuracy with 50x fewer parameters and <
  0.5 mb model size.
\newblock {\em arXiv preprint arXiv:1602.07360}, 2016.

\bibitem{landrieu2019point}
L.~Landrieu and M.~Boussaha.
\newblock Point cloud oversegmentation with graph-structured deep metric
  learning.
\newblock {\em arXiv preprint axXiv:1904.02113}, 2019.

\bibitem{landrieu2018large}
L.~Landrieu and M.~Simonovsky.
\newblock Large-scale point cloud semantic segmentation with superpoint graphs.
\newblock In {\em IEEE Conf. on Computer Vision and Pattern Recognition}, pages
  4558--4567, 2018.

\bibitem{li20173d}
B.~Li.
\newblock {3D} fully convolutional network for vehicle detection in point
  cloud.
\newblock In {\em IEEE Trans. on Intelligent Robots and Systems}, pages
  1513--1518, 2017.

\bibitem{lin2017refinenet}
G.~Lin, A.~Milan, C.~Shen, and I.~Reid.
\newblock Refinenet: Multi-path refinement networks for high-resolution
  semantic segmentation.
\newblock In {\em IEEE Conf. on Computer Vision and Pattern Recognition}, pages
  1925--1934, 2017.

\bibitem{long2015fully}
J.~Long, E.~Shelhamer, and T.~Darrell.
\newblock Fully convolutional networks for semantic segmentation.
\newblock In {\em IEEE Conf. on Computer Vision and Pattern Recognition}, pages
  3431--3440, 2015.

\bibitem{qi2017pointnet}
C.~R Qi, H.~Su, K.~Mo, and L.~J. Guibas.
\newblock Pointnet: Deep learning on point sets for {3D} classification and
  segmentation.
\newblock In {\em IEEE Conf. on Computer Vision and Pattern Recognition}, pages
  652--660, 2017.

\bibitem{ronneberger2015u}
O.~Ronneberger, P.~Fischer, and T.~Brox.
\newblock {U-net: Convolutional networks for biomedical image segmentation}.
\newblock In {\em MICCAI International Conference on Medical Image Computing
  and Computer-Assisted Intervention}, pages 234--241, 2015.

\bibitem{wang2018pointseg}
Y.~Wang, T.~Shi, P.~Yun, L.~Tai, and M.~Liu.
\newblock Pointseg: Real-time semantic segmentation based on {3D LiDAR} point
  cloud.
\newblock {\em arXiv preprint arXiv:1807.06288}, 2018.

\bibitem{wu2018squeezeseg}
B.~Wu, A.~Wan, X.~Yue, and K.~Keutzer.
\newblock Squeezeseg: Convolutional neural nets with recurrent crf for
  real-time road-object segmentation from {3D LiDAR} point cloud.
\newblock In {\em IEEE International Conference on Robotics and Automation},
  pages 1887--1893, 2018.

\bibitem{wu2018squeezesegv2}
B.~Wu, X.~Zhou, S.~Zhao, X.~Yue, and K.~Keutzer.
\newblock {SqueezesegV2}: Improved model structure and unsupervised domain
  adaptation for road-object segmentation from a {LiDAR} point cloud.
\newblock {\em arXiv preprint arXiv:1809.08495}, 2018.

\bibitem{zhao2018icnet}
H.~Zhao, X.~Qi, X.~Shen, J.~Shi, and J.~Jia.
\newblock Icnet for real-time semantic segmentation on high-resolution images.
\newblock In {\em Proc. of ECCV}, pages 405--420, 2018.

\bibitem{zhou2018voxelnet}
Y.~Zhou and O.~Tuzel.
\newblock Voxelnet: End-to-end learning for point cloud based {3D} object
  detection.
\newblock In {\em IEEE Conf. on Computer Vision and Pattern Recognition}, pages
  4490--4499, 2018.

\end{thebibliography}

\end{document}